\documentclass[runningheads]{llncs}

\usepackage[T1]{fontenc}

\usepackage{graphicx}

\usepackage{hyperref}
\usepackage{url}
\usepackage{color}

\usepackage{enumitem}
\usepackage{booktabs}

\usepackage{wrapfig}
\usepackage{graphicx}
\usepackage{multicol}
\usepackage{subcaption}
\usepackage{multibib}

\newcites{append}{Appendix bibliography}

\usepackage[capitalize]{cleveref}

\newcommand{\notw}{NO\textsubscript{2}}

\begin{document}

\title{Bayesian Optimisation Against Climate Change: \\ Applications and Benchmarks}

\titlerunning{Bayesian Optimisation Against Climate Change}

\author{Sigrid Passano Hellan\inst{1}\orcidID{0009-0002-6443-0395} \and
Christopher G. Lucas\inst{2}\orcidID{0000-0002-6655-8627} \and
Nigel H. Goddard\inst{2}\orcidID{0000-0002-1731-3412}
}

\authorrunning{S. P. Hellan et al.}

\institute{ NORCE Norwegian Research Centre 
\email{sipa@norceresearch.no} \\ and
Bjerknes Centre for Climate Research, Bergen, Norway \and
School of Informatics, University of Edinburgh, UK
}

\maketitle

\begin{abstract}

Bayesian optimisation is a powerful method for optimising black-box functions, popular in settings where the true function is expensive to evaluate and no gradient information is available. 
    Bayesian optimisation can improve responses to many optimisation problems within climate change for which simulator models are unavailable or expensive to sample from. While there have been several demonstrations of climate-related applications, there has been no unifying review of applications and benchmarks. We provide such a review here, to encourage the use of Bayesian optimisation for important and well-suited applications.
    We identify four main application domains: material discovery, wind farm layout, optimal renewable control and environmental monitoring.
    For each domain we
    identify a public benchmark or data set that is easy to use and evaluate systems against, while being representative of real-world problems.
    Due to the lack of a suitable benchmark for environmental monitoring, we propose LAQN-BO, based on air pollution data.
    Our contributions are: a) summarising Bayesian optimisation applications related to climate change;
    b) identifying a representative range of benchmarks, providing example code where necessary; and c) introducing a new benchmark, LAQN-BO.

\end{abstract}
 
\section{Introduction}

The use of machine learning (ML) to tackle climate change is gaining traction, including ML-wide surveys of relevant applications \cite{rolnick2022tackling,donti2021machine}.
To facilitate technical innovation and its adoption into practice, it is important to complement these reviews with concise summaries of problems that specific ML frameworks are well-matched to, along with example data sets. Here, we provide a summary, paired with data sets, for Bayesian optimisation.
This is important for three reasons: 
a) it reduces start-up costs and overhead, as the time-consuming task of identifying suitable data sets does not have to be repeated; 
b) it helps researchers focus on problems that are timely and/or important, and have a structure that is suitable for the framework(s) they are using;
and c) 
it makes it easier to compare methods, as numerical results can be compared directly. 
The benchmarks should be similar to real-world applications. That will provide more realistic expectations of performance and 
show which versions of Bayesian optimisation are most promising, e.g. in terms of underlying models. 
It will also identify remaining challenges, e.g. in terms of generating useful priors from related data. 

\renewcommand{\theenumi}{\alph{enumi}}

Bayesian optimisation (BO), is a machine learning surrogate-based approach to black-box optimisation \cite{garnett2023bayesian}. 
It is particularly suited to problems that: 
\begin{enumerate}[noitemsep,topsep=0pt]
    \item Have a complex unknown structure which can only be efficiently modelled through a surrogate model, e.g. the conductivity of solar panel materials.
    \item Are expensive or slow to evaluate, requiring sample efficiency from trying out fewer bad or uninformative choices, e.g. when synthesising new materials. 
    \item Require identifying extrema, either for optimisation, e.g. maximising power generation, or for intervention, e.g. emission monitoring. 
\end{enumerate}

\begin{figure}
\centering
 \includegraphics[width=0.5\linewidth]{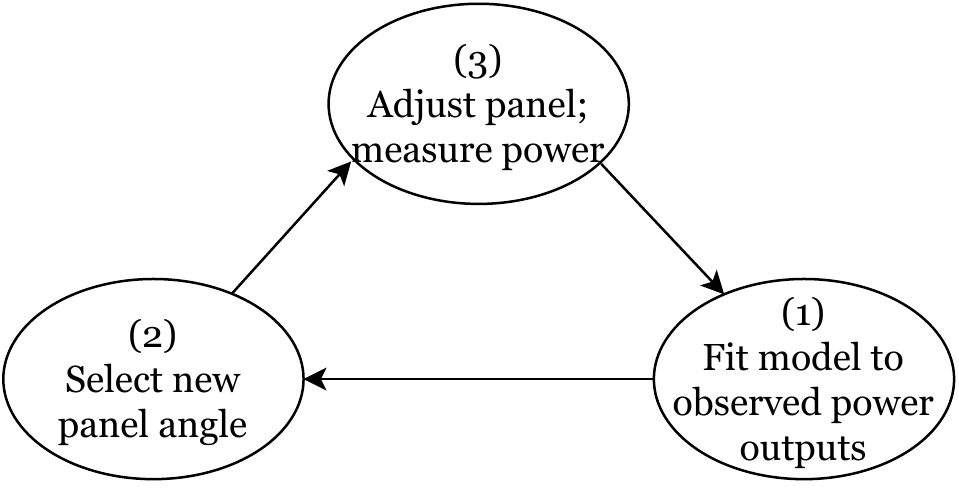}
 \caption{Bayesian optimisation to maximise solar panel power generation. }
\label{fig:bo-overview}
\end{figure}

Bayesian optimisation consists of fitting a surrogate model, typically a Gaussian process \cite{rasmussen_gaussian_2006}, to observed data, and using the surrogate model to choose what input value to evaluate next. Then the surrogate model is adjusted, another input value evaluated, and so on. 
For example, Bayesian optimisation can be used to find the angle of a solar panel that maximises power output, see \cref{fig:bo-overview}: (1)~model the output as a function of angle based on all available data and/or domain knowledge; (2) select a promising angle to try next, e.g., an angle that is a good candidate for maximising power output or that is maximally informative about the best angle; and (3) try this angle before updating the model and returning to Step~1.
An additional benefit of Bayesian optimisation is that it explicitly models its uncertainty, which can be used to determine when to stop optimising \cite{makarova2022automatic}, or to trade off exploration and exploitation \cite{de2021greed}.  

We present a summary of Bayesian optimisation as applied to climate change. We identify the main application domains, emphasise their climate-related impact, and identify beginner-friendly benchmarks for each. We found a benchmark to be lacking for environmental monitoring, so provide LAQN-BO, based on air pollution data \cite{imperial_college_london_london_1993_url}. 
The four use cases are:
\textit{material discovery} --- accelerating the development of e.g. new solar panels; \textit{wind farm layouts} --- choosing where to place individual turbines; \textit{optimal renewable control} --- choosing operating parameters of solar panels or wind turbines; and \textit{environmental monitoring} --- choosing where to place sensors. 
In this paper we mainly focus on Bayesian optimisation. However, there are other closely related methods
such as Bayesian experimental design 
\cite{rainforth2023modern,valentin2023designing} 
and
we have opted to include some examples of these methods as well, e.g. \cite{tran2018active,kleinegesse2020bayesian}.
\cite{bliek2022survey} is a related survey of Bayesian optimisation and other surrogate-based optimisation methods. Where \cite{bliek2022survey} takes a higher-level approach, focusing on  methodology, we instead group work by applications in order to identify data sets and applications for new projects.
When selecting benchmarks, we have emphasised ease of adoption and open access. 

Bayesian optimisation is not a complete solution, but an important tool when combined with domain knowledge, and can help improve responses to climate change.
Currently, Bayesian optimisation is often evaluated on synthetic functions or hyperparameter optimisation (HPO) benchmarks.
But performance in real applications is the ultimate test of any ML method, including Bayesian optimisation. HPO is just one kind of problem and synthetic problems often fail to anticipate crucial challenges, e.g. in terms of finding suitable inductive biases
\cite{hellan_bayesian_2022}.
To increase the impact of BO, and its prevalence, we should adopt a broader set of benchmarks. 
That would not only demonstrate which methods work well in practice, but open up new research problems to solve.
Using benchmarks related to climate change and sustainability has the advantage of doing so in an innovative field, increasing the chance of adoption, and with real impact.

\section{Material discovery}
Material discovery means developing materials with superior combinations of properties \cite{frazier2016bayesian}, e.g. more efficient solar cells.
Bayesian optimisation and related methods have been suggested for optimising 
a wide range of materials, from concrete \cite{severson2021amortized,ament2023sustainable} and direct air capture of CO\textsubscript{2} \cite{ortiz-montalvo2021machine} to solar panel glass \cite{haghanifar2019using}, electrifying  transport and chemical industries \cite{annevelink2022automat}, and for 
electrocatalysts for CO\textsubscript{2} reduction and H\textsubscript{2} production \cite{tran2018active,zhong2020accelerated,frey_neyerlin_modestino_2022}. 
The benefits are in reducing greenhouse gas emissions or increasing their capture;
making renewable energy generation or storage more efficient; or running more sectors off renewable energy.
If evaluating the material requires synthesising it, there are huge potential gains in time and cost by only doing so on the most promising candidates. 

\cite{liang2021benchmarking} provide three material discovery data sets \cite{sun2021data,bash2021multi,mekki2021two} related to solar panels, with 3--5 features each and 94--178 unique evaluated data points. Using \cref{fig:bo-overview} to illustrate the Bayesian optimisation process, the steps become: (1) model the material performance as a function of ingredient proportions used; (2) select a new combination of ingredient proportions to try; and (3) produce and evaluate the material. We replace (3) with looing up the value in the benchmark.

\begin{itemize}[noitemsep]
    \item \textbf{Benchmark:}  \url{https://github.com/PV-Lab/Benchmarking} \cite{liang2021benchmarking}
    \item \textbf{Features:} Material properties: 3--5 dimensions
    \item \textbf{Data type:} CSV files
    \item \textbf{Sampling:} Manufacture material. In benchmark replaced by table look-up
    \item \textbf{Objective:} Material performance: conductivity, absorbance or stability
    \item \textbf{Impact:} Better solar panels; more renewables and less climate gases
\end{itemize}

\section{Wind farm layout}

\begin{figure}[h]
\centering
\begin{subfigure}{.99\textwidth}
    \centering
    \includegraphics[width=.99\linewidth]{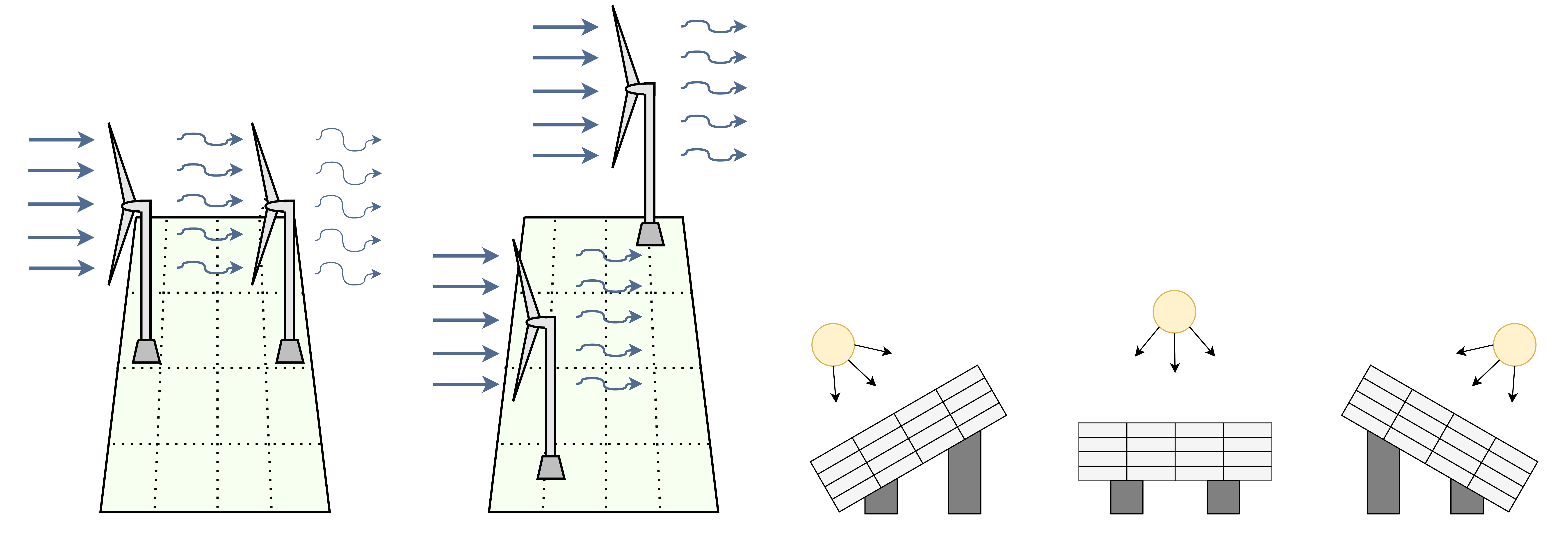}  
    \caption{\textbf{Left:} Planning of wind turbine layouts within limits (green). The wind is shown by blue arrows. Second from left: both turbines get unhindered wind and will produce more electricity. \textbf{Right:} Adjusting solar panels to optimise power generation.}
    \label{fig:wind-solar}
\end{subfigure}
\begin{subfigure}{.99\textwidth}
    \centering
    \vspace{0.15cm}
    \includegraphics[width=.99\linewidth]{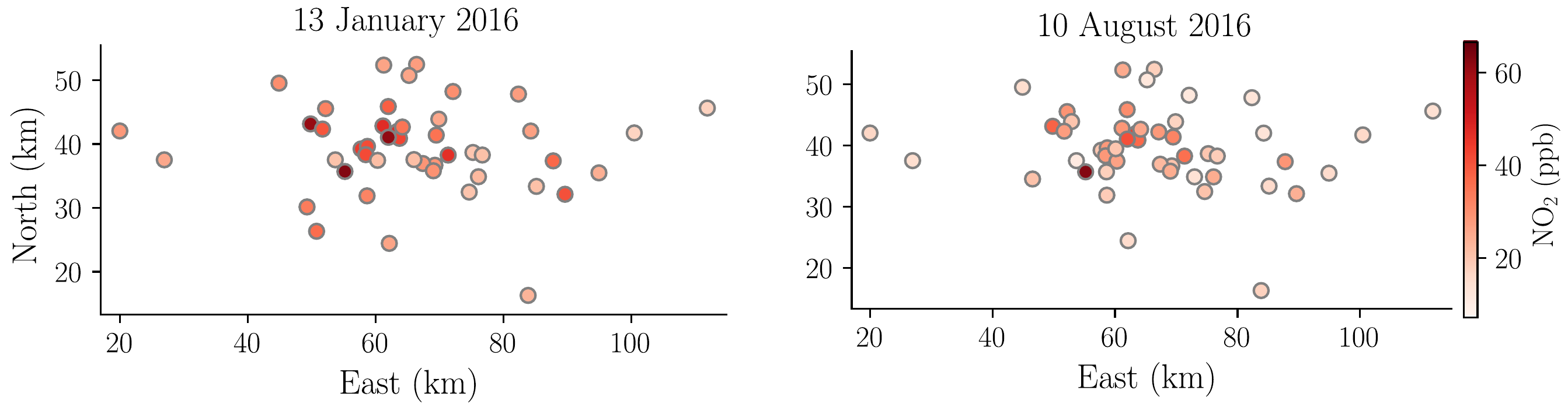}  
    \caption{Example environmental measurements.}
    \label{fig:env-examples}
\end{subfigure}
\caption{Benchmark illustrations. For detailed explanations see \cref{sec-app-illust}.}
\label{fig:benchmarks-together}
\end{figure}

Wind turbines are typically grouped together in wind farms, with several turbines in relatively close proximity, 
to reduce
installation costs and environmental impacts.
Before construction, the locations of individual wind turbines must be planned. 
Wind turbines work by extracting kinetic energy from the air, so the wind is weaker and more turbulent after flowing past a turbine. This leads to less power generation \cite{park2016bayesian} and greater dynamic loads and fatigue on the downwind turbines \cite{dong2022wind}. This is illustrated in \cref{fig:wind-solar} (left), where the two turbines need to be placed within the green area and the wind is shown by arrows.
Determining the optimal wind turbine layout is a difficult but important optimisation problem, since more renewable electricity can be generated without requiring more turbines to be installed. 
The efficiency of Bayesian optimisation is important for this problem, since the type of simulations used can take 15 seconds to run on one CPU even for five wind turbines \cite{bliek2021expobench}, and the simulations must be run many times to compare layouts.
Bayesian optimisation can be applied to the problem by replacing the steps in \cref{fig:bo-overview} with (1) modelling the collective power output as a function of the wind turbine placements; (2) selecting a new combination of locations to evaluate; and (3) running the power output simulation.

\cite{bliek2021expobench}'s 
set of benchmarking problems and baselines include a wind farm layout problem, \textit{windwake}, based on FLORIS \cite{FLORIS_2020} simulations. 
The benchmark assumes a fixed number of wind turbines; this is relaxed in \cite{chugh2022wind}, where the two objectives of cost and power generation are jointly optimised, using simulations based on \cite{pedersen2019dtuwindenergy}.
In related pieces of work, \cite{mern2021bayesian} build wind maps for later layout planning, and \cite{tillmann2020minimising} plan the layout of
bifacial solar panel arrays. We use the benchmark in \cite{bliek2021expobench} as it has been been prepared for further use.

\begin{itemize}[noitemsep]
    \item \textbf{Benchmark:}  \url{https://github.com/AlgTUDelft/ExpensiveOptimBenchmark} 
    \item \textbf{Features:} Spatial locations: 10 dimensions. 2 for each of 5 turbines \cite{bliek2021expobench}
    \item \textbf{Data type:} Simulation
    \item \textbf{Sampling:} Simulate layout
    \item \textbf{Objective:} Energy production
    \item \textbf{Impact:} More renewable generation, which reduces climate gas emissions
\end{itemize}

\section{Optimal renewable control}

Having planned the layout for a renewable power farm, the next question is how best to operate it \cite{dong2022wind}. For wind turbines the blade angles \cite{doekemeijer2019model,park2016bayesian,yang2022cooperative}, and induction factor \cite{park2016bayesian} need to be set, and for airborne systems the altitude \cite{ramesh2022movement}. 
For solar panels the voltage level applied \cite{abdelrahman2016bayesian,lyden2018bayesian} and the angle of the panel \cite{abel2018bandit} can be adjusted.  
By adjusting these to maximise energy generation, we minimise emissions for a given electricity demand. 
For wind farms, the optimisation should occur jointly, as the available wind energy for a downstream turbine is impacted by the operating settings of the upstream turbine \cite{park2016bayesian}. 
With Bayesian optimisation the response of the non-linear problem can be learnt through data collection.
Extensive work on wind turbine control has been done using Bayesian Ascent, a version of BO which limits changes in inputs between iterations, see e.g.
\cite{park2016bayesian}.
Relatedly, \cite{moustakis2019practical} tune the parameters of traditional control methods, and \cite{mulders2020efficient} reduce blade fatigue.
\cite{andersson2020real} use a related method for wind turbine control.

\cite{abel2018bandit} provide a benchmark for solar panel optimisation. 
The goal is to maximise  energy yield by adjusting  solar panel angles, see \cref{fig:wind-solar} (right). They evaluate contextual bandits and reinforcement learning methods, but the problem is also suitable for BO: we provide a basic example at 
\url{https://github.com/sighellan/solar_panels_rl}, see \cref{sec-app-solar_rl}. \cref{fig:bo-overview} shows how BO is applied to the problem. We model power output as a function of panel angles; decide on a new angle to try; and run the simulation to evaluate the new angle.
The challenge is not tracking the sun; its location is predictable and given as an input. Instead, the idea is to adapt to other changes such as cloud cover. 
However, the angle of many solar panels cannot be adjusted.
Wind turbines have more control settings, interact in less predictable ways, and have seen more applications of BO. We chose the benchmark in \cite{abel2018bandit} because it is well-documented and publicly available, but real-world benefits are more likely to come from applications to wind turbines.

\begin{itemize}[noitemsep]
    \item \textbf{Benchmark:} \url{https://github.com/david-abel/solar_panels_rl} \cite{abel2018bandit}
    \item \textbf{Features:} Panel direction: 2 dimensions
    \item \textbf{Data type:} Simulation
    \item \textbf{Sampling:} Simulate collected sunlight
    \item \textbf{Objective:} Collected energy
    \item \textbf{Impact:} More renewable generation, which reduces climate gas emissions
\end{itemize}

\section{Environmental monitoring}

An important problem within environmental monitoring is that of choosing where to place sensors. 
This is difficult since the planning must be done without knowing the usefulness of each site, which depends on future data.
BO and related methods are useful tools 
in this setting, as they build probabilistic models of the environmental characteristic of interest, e.g. air pollution,
which they use to make efficient choices.
If we are mainly interested in the worst-hit locations, e.g. for targeting interventions or evaluating legal compliance, BO lets us do that efficiently. 
To build accurate environmental maps for an entire area we can instead use Bayesian experimental design, e.g. if we want to know the water quality throughout a lake. 
Monitoring is important for tracking the impacts of climate change, including on oceans \cite{sanchez2020artificial}, for locating gas leakages \cite{asenov_active_2019,gao2022bayesian,kleinegesse2020bayesian}, and the related task of air pollution monitoring \cite{ainslie_application_2009,morere_sequential_2017,marchant_bayesian_2012,singh_modeling_2010,hellan_optimising_2020_arxiv,hellan_bayesian_2022,leu2020air}. 
\cite{samaniego2021bayesian} provide a benchmark for path-planning based on water quality monitoring, also used by \cite{folch2022snake}, but the benchmark uses synthetic data.

\textbf{The LAQN-BO benchmark:} We present a new climate change related benchmark, using air pollution data from the London Air Quality Network (LAQN) \cite{imperial_college_london_london_1993_url}, see examples in \cref{fig:env-examples}..
Air pollution is related to climate change in that the sources and hence required interventions are often interlinked, e.g. car exhaust. Also, environmental changes are often sped up by climate change, making their monitoring more important.
The optimisation problems in the LAQN-BO benchmark are similar to those in \cite{hellan_bayesian_2022}. But while that focuses on methodology, we here focus on the problems themselves. We provide simple scripts and step-by-step instructions for generating the problems and put them in the context of other climate-related BO problems and benchmarks. 

The optimisation objective is the \notw{} concentration, i.e. to find the location of maximum pollution from the set of available locations. This is useful as higher pollution concentrations have greater impacts on human health.
Using \cref{fig:bo-overview} as a basis, the steps become: (1) model the \notw{} concentration as a function of spatial location; (2) determine a new location to evaluate; and (3) measure the pollution in the new location. We use LAQN-BO to replace (3) with a historical measurement.
We construct training and test sets from the 2015 and 2016 data, respectively. Each problem corresponds to a single day, so multiple days give multiple problems. As in \cite{hellan_bayesian_2022}, we filter out days when fewer than 40 stations collected measurements, and only use the `Roadside' sensors, resulting in 214 training problems and 365 test problems.
We log-transformed the data, and standardised using the mean and standard deviation from the training set.

The LAQN-BO benchmark has several advantages. Firstly, it uses real data. Secondly, we present a \textit{set} of related problems, so that methods can be trained on the training set, and evaluated on the test set. Finally, the provided scripts can be easily adapted to generate new problems using other pollutants or  years of the extensive  LAQN data. None of the other benchmarks combine all of these advantages. A disadvantage is that data are not available everywhere, only at locations with existing sensors, limiting the available search space. 
This corresponds to having to choose from a sparse set of locations. For instance, if we are mounting sensors to lampposts, we can only choose between existing lampposts.

\begin{itemize}[noitemsep]
    \item \textbf{Benchmark:} \url{https://github.com/sighellan/LAQN-BO} 
    \item \textbf{Features:} Spatial coordinates: 2 dimensions
    \item \textbf{Data type:} Python class
    \item \textbf{Sampling:} Place pollution sensor. In benchmark replaced by table look-up
    \item \textbf{Objective:} \notw{} concentration
    \item \textbf{Impact:} Better monitoring of air pollution, enabling targeted interventions
\end{itemize}

\section{Benchmark comparison}

The four benchmarks identified in our survey have different characteristics, as is summarised in \cref{tab:benchmark-comparison}. 
They represent different key challenges for Bayesian optimisation deployments, and provide breadth for evaluating different types of Bayesian optimisation methods.
 The key challenge of the materials benchmark is its small size, meaning there is little opportunity to fine-tune the methods. 
For LAQN-BO it is that of learning from the training data, as the problems have as little as 40 evaluations each, requiring very sample-efficient learning.
The wind farm layout benchmark also requires efficient use of data. But in contrast to LAQN-BO, it does not provide training data for transfer learning.
The key challenge for the renewable control benchmark is dealing with large numbers of samples, as the panel direction is optimised regularly over many days. It is therefore a good candidate for testing Bayesian optimisation built on scalable Gaussian processes \cite{liu2020gaussian}. 
It also requires keeping track of context, e.g. time, as the sun moves across the sky. 
An additional contrast to the other benchmarks is that the performance throughout the optimisation matters, as it directly impacts the amount of electricity generated.

\begin{table}[h]
\caption{Comparison of benchmarks.}
\label{tab:benchmark-comparison}
\begin{tabular}{@{}lllrr@{}}
\toprule
Benchmark                & Type       & Key challenge             & Dimensionality         &\# problems \\ \midrule
Materials \cite{liang2021benchmarking}               & Real       & Small size                 & 3/5               & 3                               \\
Wind farm layout \cite{bliek2021expobench}        & Simulation & Sample efficiency         & 10               & 1                               \\
Renewable control \cite{abel2018bandit}       & Simulation & Scaling, context & 2               & 10                              \\
Environmental monitoring  & Real       & Transfer learning                    & 2 & 214+365                         \\ \bottomrule
\end{tabular}
\end{table}

The materials benchmark is the easiest to start using, as the data are provided in CSV files, followed by that for environmental monitoring, LAQN-BO, which provides a simple Python interface to the data. The benchmarks for wind farm layout and renewable control require more setup, as methods need to be interfaced with the simulators. Their advantage is that they are easier to extend, and can be evaluated for more input values.
LAQN-BO can be easily extended to more problems using the LAQN data \cite{imperial_college_london_london_1993_url}, but extensions to other base data sets would require more work. The materials benchmark is the hardest to extend, as it relies on results from synthesising and evaluating materials.

\section{Conclusion and other applications}

There are many more climate-related applications: learning energy consumptions of  household appliances \cite{jia2019active};  optimising charging protocols for electric vehicles  \cite{attia2020closed}; scheduling smart appliances to smooth out demand curves \cite{tabakhi2020smart}; improving simulations \cite{rodriguez2024enhancing} and experiments \cite{gundecha2024metalearned} for nuclear fusion; reducing vibrations for floating wind turbines \cite{zhang2022vibration}; reducing the impact of livestock diseases made more prevalent by global warming \cite{spooner2020bayesian}; and
tuning parameters for heating and cooling system and building models to reduce energy demands \cite{zhan2022calibrating,chakrabarty2021scalable,fiducioso2019safe}.
Bayesian optimisation also has applications in improving climate models,
by targeting informative training data \cite{watson2021machine}, and in polar climate science \cite{bente2022probabilistic}. And it has been used for parameter tuning of a global land surface model \cite{druel2017towards}, a landslide model \cite{pradhan2021meta} and forecasting models for electricity generation and demand \cite{trivedi2021peak}.

To encourage more Bayesian optimisation practitioners to work on climate change applications, it is important to establish more benchmarks, standard data sets or simulators, as we do here with the LAQN-BO benchmark. 
More challenging benchmarks should also be established, e.g. for constrained or multi-objective optimisation, to encompass more fully the complexity of real problems.
For example: some of the area might be unsuitable for building wind turbines; and when building generators there are multiple competing objectives \cite{wu2018efficiently,flecker2022reducing}. 
Additionally, there are constraints for controlling the generators, due to increased energy costs and fatigue when changing settings. 
The energy cost of changing angles is included in the renewable control benchmark \cite{abel2018bandit}, so Bayesian Ascent \cite{park2016bayesian} should be tested, as it limits the difference between iterations.
As a further extension, the layout and control of wind farms can be optimised jointly \cite{chen2022joint}.

In the future, all these benchmarks should be brought together and their interfaces aligned, to create a standardised, climate-themed benchmark suite.
By identifying four benchmarks representing the main application domains, introducing a new benchmark, and highlighting the potential for Bayesian optimisation, we take an important step in facilitating future research.

\begin{credits}
\subsubsection{\ackname}
We thank Linus Ericsson, Simon Valentin and Michael Gutmann for helpful suggestions.
This work was supported by the EPSRC Centre for Doctoral Training in Data Science, funded by the UK Engineering and Physical Sciences Research Council (grant EP/L016427/1) and the University of Edinburgh.

\subsubsection{\discintname}
The authors have no competing interests to declare that are
relevant to the content of this article. 
\end{credits}

\bibliographystyle{splncs04}
\bibliography{short}

\begin{thebibliography}{1}
\providecommand{\url}[1]{\texttt{#1}}
\providecommand{\urlprefix}{URL }
\providecommand{\doi}[1]{https://doi.org/#1}

\bibitem{abel2019simple_rl}
Abel, D.: simple\_rl: Reproducible reinforcement learning in python. ICLR Workshop on Reproducibility in Machine Learning  (2019)

\bibitem{balandat2020botorch}
Balandat, M., et~al.: {BoTorch: A Framework for Efficient Monte-Carlo Bayesian Optimization}. In: NeurIPS (2020)

\end{thebibliography}


\begin{thebibliography}{10}
\providecommand{\url}[1]{\texttt{#1}}
\providecommand{\urlprefix}{URL }
\providecommand{\doi}[1]{https://doi.org/#1}

\bibitem{abdelrahman2016bayesian}
Abdelrahman, H., et~al.: Bayesian optimization for maximum power point tracking in photovoltaic power plants. European Control Conference  (2016)

\bibitem{abel2018bandit}
Abel, D., et~al.: Bandit-based solar panel control. In: AAAI (2018)

\bibitem{ainslie_application_2009}
Ainslie, B., et~al.: Application of an entropy-based {Bayesian} opt. [...] monitoring network for single air pollutants. Env. Management  (2009)

\bibitem{ament2023sustainable}
Ament, S., Witte, A.C., Garg, N., Kusuma, J.: Sustainable concrete via {Bayesian} optimization. NeurIPS 2023 RealML Workshop  (2023)

\bibitem{andersson2020real}
Andersson, L.E., Imsland, L.: Real-time optimization of wind farms using modifier adaptation and machine learning. Wind Energy Science  (2020)

\bibitem{annevelink2022automat}
Annevelink, E., et~al.: Automat: Automated materials discovery for electrochemical systems. MRS Bulletin  (2022)

\bibitem{asenov_active_2019}
Asenov, M., et~al.: Active {Localization} of {Gas} {Leaks} {Using} {Fluid} {Simulation}. IEEE Robotics and Automation Letters  (2019)

\bibitem{attia2020closed}
Attia, P.M., et~al.: Closed-loop optimization of fast-charging protocols for batteries with machine learning. Nature  (2020)

\bibitem{bash2021multi}
Bash, D., et~al.: Multi-fidelity high-throughput optimization of elect. conductivity in {P3HT-CNT} composites. Advanced Functional Materials  (2021)

\bibitem{bente2022probabilistic}
Bente, K., et~al.: Probabilistic {ML} in polar earth and climate science  (2022)

\bibitem{bliek2022survey}
Bliek, L.: A survey on sustainable surrogate-based opt. Sustainability  (2022)

\bibitem{bliek2021expobench}
Bliek, L., et~al.: Benchmarking surrogate-based optimisation algorithms on expensive black-box functions. Applied Soft Computing  (2023)

\bibitem{chakrabarty2021scalable}
Chakrabarty, A., et~al.: Scalable {B}ayesian optimization for model calibration. Energy and Buildings  (2021)

\bibitem{chen2022joint}
Chen, K., Lin, J., Qiu, Y., Liu, F., Song, Y.: Joint optimization of wind farm layout considering optimal control. Renewable Energy  (2022)

\bibitem{chugh2022wind}
Chugh, T., Ymeraj, E.: Wind farm layout optimisation using set based multi-objective {Bayesian} optimisation. In: GECCO (2022)

\bibitem{de2021greed}
De~Ath, G., et~al.: Greed is good: Exploration and exploitation trade-offs in {Bayesian} opt. Evolutionary Learning and Optimization ACM  (2021)

\bibitem{doekemeijer2019model}
Doekemeijer, B.M., et~al.: Model-based closed-loop wind farm control for power maximization using {B}ayesian optimization. In: IEEE CCTA (2019)

\bibitem{dong2022wind}
Dong, H., Xie, J., Zhao, X.: Wind farm control technologies: From classical control to reinforcement learning. Progress in Energy  (2022)

\bibitem{donti2021machine}
Donti, P.L., Kolter, J.Z.: Machine learning for sustainable energy systems. Annual Review of Environment and Resources  (2021)

\bibitem{druel2017towards}
Druel, A., et~al.: Towards a more detailed representation of high-latitude vegetation in [...] {ORCHIDEE}. Geoscientific Model Development  (2017)

\bibitem{fiducioso2019safe}
Fiducioso, M., et~al.: Safe contextual {Bayesian} optimization for sustainable room temperature {PID} control tuning. In: IJCAI (2019)

\bibitem{flecker2022reducing}
Flecker, A.S., et~al.: Reducing adverse impacts of {Amazon} hydropower expansion. Science  (2022)

\bibitem{folch2022snake}
Folch, J.P., et~al.: {SnAKe}: {Bayesian} optimization with pathwise exploration. NeurIPS  (2022)

\bibitem{frazier2016bayesian}
Frazier, P.I., Wang, J.: Bayesian optimization for materials design. In: Information science for materials discovery and design. Springer (2016)

\bibitem{frey_neyerlin_modestino_2022}
Frey, D., et~al.: Bayesian optimization of electrochemical devices for electrons-to-molecules conversions. ChemRxiv  (2022)

\bibitem{gao2022bayesian}
Gao, T., Bai, X.: Bayesian opt.-based three-dimensional, time-varying env. monitoring using an {UAV}. Journal of Intelligent \& Robotic Systems  (2022)

\bibitem{garnett2023bayesian}
Garnett, R.: Bayesian optimization. Cambridge University Press (2023)

\bibitem{gundecha2024metalearned}
Gundecha, V., et~al.: Meta-learned {Bayesian} optimization for energy yield in inertial confinement fusion. NeurIPS CCAI Workshop  (2024)

\bibitem{haghanifar2019using}
Haghanifar, S., et~al.: Using {Bayesian} optimization to improve solar panel performance by [...]. ICML CCAI Workshop  (2019)

\bibitem{hellan_optimising_2020_arxiv}
Hellan, S.P., Lucas, C.G., Goddard, N.H.: Optimising placement of pollution sensors in windy environments. NeurIPS AI4Earth Workshop  (2020)

\bibitem{hellan_bayesian_2022}
Hellan, S.P., Lucas, C.G., Goddard, N.H.: Bayesian optimisation for active monitoring of air pollution. AAAI  (2022)

\bibitem{imperial_college_london_london_1993_url}
Imperial College~London, E.R.G.: London {Air} {Quality} {Network} (1993)

\bibitem{jia2019active}
Jia, Y., et~al.: Active collaborative sensing for energy breakdown. In: ACM Information and Knowledge Management (2019)

\bibitem{kleinegesse2020bayesian}
Kleinegesse, S., Gutmann, M.U.: Bayesian experimental design for implicit models by mutual information neural estimation. In: ICML (2020)

\bibitem{leu2020air}
Leu, F.Y., Ho, J.S.: Air pollution source identification by using neural network with {Bayesian} optimization. IMIS  (2020)

\bibitem{liang2021benchmarking}
Liang, Q., et~al.: Benchmarking the performance of {B}ayesian opt. across multiple exp. materials science domains. npj Comp. Materials  (2021)

\bibitem{liu2020gaussian}
Liu, H., et~al.: When {Gaussian} process meets big data: A review of scalable {GPs}. IEEE neural networks and learning systems  (2020)

\bibitem{lyden2018bayesian}
Lyden, S., Olding, W., Darbari, Z.: Bayesian optimization for maximum power point tracking in photovoltaic systems. In: PESGM. IEEE (2018)

\bibitem{makarova2022automatic}
Makarova, A., et~al.: Automatic termination for hyperparameter optimization. AutoML  (2022)

\bibitem{marchant_bayesian_2012}
Marchant, R., Ramos, F.: Bayesian optimisation for {Intelligent} {Environmental} {Monitoring}. In: {Intelligent} {Robots} and {Systems}. IEEE (2012)

\bibitem{mekki2021two}
Mekki-Berrada, F., et~al.: Two-step machine learning enables optimized nanoparticle synthesis. npj Computational Materials  (2021)

\bibitem{mern2021bayesian}
Mern, J., Yildiz, A., Sunberg, Z., Mukerji, T., Kochenderfer, M.J.: Bayesian optimized {Monte} {Carlo} planning. In: AAAI (2021)

\bibitem{morere_sequential_2017}
Morere, P., et~al.: Sequential {Bayesian} optimization as a {POMDP} for environment monitoring with {UAVs}. In: ICRA. IEEE (2017)

\bibitem{moustakis2019practical}
Moustakis, N., et~al.: A practical {Bayesian} optimization approach for the optimal estimation of the rotor effective wind speed. In: ACC. IEEE (2019)

\bibitem{mulders2020efficient}
Mulders, S., et~al.: Efficient tuning of individual pitch control. In: Journal of Physics: Conference Series. IOP Publishing (2020)

\bibitem{FLORIS_2020}
NREL: {FLORIS}. version 2.1.1 (2020), \url{https://github.com/NREL/floris}

\bibitem{ortiz-montalvo2021machine}
Ortiz-Montalvo, D.L., et~al.: Machine learning for climate change: [...] for direct air capture of {CO2}. In: ICML CCAI Workshop (2021)

\bibitem{park2016bayesian}
Park, J., Law, K.H.: Bayesian ascent: A data-driven optimization scheme for real-time control [...]. IEEE Control Systems Technology  (2016)

\bibitem{pedersen2019dtuwindenergy}
Pedersen, M., et~al.: Dtuwindenergy/pywake: Pywake. Zenodo [code]  (2019)

\bibitem{pradhan2021meta}
Pradhan, B., et~al.: A meta-learning approach of optimisation for spatial prediction of landslides. Remote Sensing  (2021)

\bibitem{rainforth2023modern}
Rainforth, T., et~al.: Modern {Bayesian} exp. design. Statistical Science  (2024)

\bibitem{ramesh2022movement}
Ramesh, S.S., et~al.: Movement penalized {Bayesian} optimization with application to wind energy systems. NeurIPS  (2022)

\bibitem{rasmussen_gaussian_2006}
Rasmussen, C.E., Williams, C.K.I.: Gaussian Processes for ML. MIT (2006)

\bibitem{rodriguez2024enhancing}
Rodriguez-Fernandez, P., et~al.: Enhancing predictive capabilities in fusion burning plasmas through surrogate-based opt. [...]. Nuclear Fusion  (2024)

\bibitem{rolnick2022tackling}
Rolnick, D., et~al.: Tackling climate change with machine learning. ACM Computing Surveys (CSUR)  (2022)

\bibitem{samaniego2021bayesian}
Samaniego, F.P., et~al.: A {Bayesian} opt. approach for water resources monitoring through an autonomous surface vehicle. IEEE Access  (2021)

\bibitem{sanchez2020artificial}
Sanchez-Pi, N., et~al.: Artificial intelligence, machine learning and modeling [...] oceans and climate change. In: NeurIPS CCAI Workshop (2020)

\bibitem{severson2021amortized}
Severson, K., et~al.: Amortized inference of {G}aussian process hyperparameters for improved concrete strength [...]. NeurIPS CCAI workshop  (2021)

\bibitem{singh_modeling_2010}
Singh, A., et~al.: Modeling and decision making in spatio-temporal processes for environmental surveillance. In: {IEEE} {Robotics} and {Automation} (2010)

\bibitem{spooner2020bayesian}
Spooner, T., et~al.: Bayesian optimisation of restriction zones for bluetongue control. Scientific reports  (2020)

\bibitem{sun2021data}
Sun, S., et~al.: A data fusion approach to optimize compositional stability of halide perovskites. Matter  (2021)

\bibitem{tabakhi2020smart}
Tabakhi, A., et~al.: The smart appliance scheduling problem: {A} {B}ayesian optimization approach. Principles and Practice of Multi-Agent Systems  (2020)

\bibitem{tillmann2020minimising}
Tillmann, P., et~al.: Minimising the levelised cost of electricity for bifacial solar panel arrays using {B}ayesian opt. Sustainable Energy \& Fuels  (2020)

\bibitem{tran2018active}
Tran, K., Ulissi, Z.W.: Active learning across intermetallics [...] of electrocatalysts for {CO2} reduction and {H2} evolution. Nature Catalysis  (2018)

\bibitem{trivedi2021peak}
Trivedi, R., Khadem, S.: Peak demand management and schedule optimisation for energy storage [...]. EUROCON IEEE Smart Technologies  (2021)

\bibitem{valentin2023designing}
Valentin, S., et~al.: Designing optimal behavioral exp. using {ML}. Elife  (2024)

\bibitem{watson2021machine}
Watson-Parris, D.: Machine learning for weather and climate are worlds apart. Philosophical Transactions of the Royal Society A  (2021)

\bibitem{wu2018efficiently}
Wu, X., et~al.: Efficiently approximating the pareto frontier: hydropower dam placement in the {Amazon} basin. In: AAAI (2018)

\bibitem{yang2022cooperative}
Yang, S., et~al.: Cooperative yaw control of wind farm using a double-layer machine learning framework. Renewable Energy  (2022)

\bibitem{zhan2022calibrating}
Zhan, S., et~al.: Calibrating building simulation models using multi-source datasets and meta-learned {B}ayesian opt. Energy and Buildings  (2022)

\bibitem{zhang2022vibration}
Zhang, X., et~al.: Vibration suppression of multi-component floating structures via passive {TMDs} and {Bayesian} ascent. Ocean Engineering  (2022)

\bibitem{zhong2020accelerated}
Zhong, M., et~al.: Accelerated discovery of {CO2} electrocatalysts using active machine learning. Nature  (2020)

\end{thebibliography}

\newpage

\appendix

\section{Benchmark illustrations} \label{sec-app-illust}

The benchmark illustrations in \cref{fig:benchmarks-together} are reproduced here with more detailed descriptions.

\begin{figure}[h]
\centering
 \includegraphics[width=0.8\linewidth]{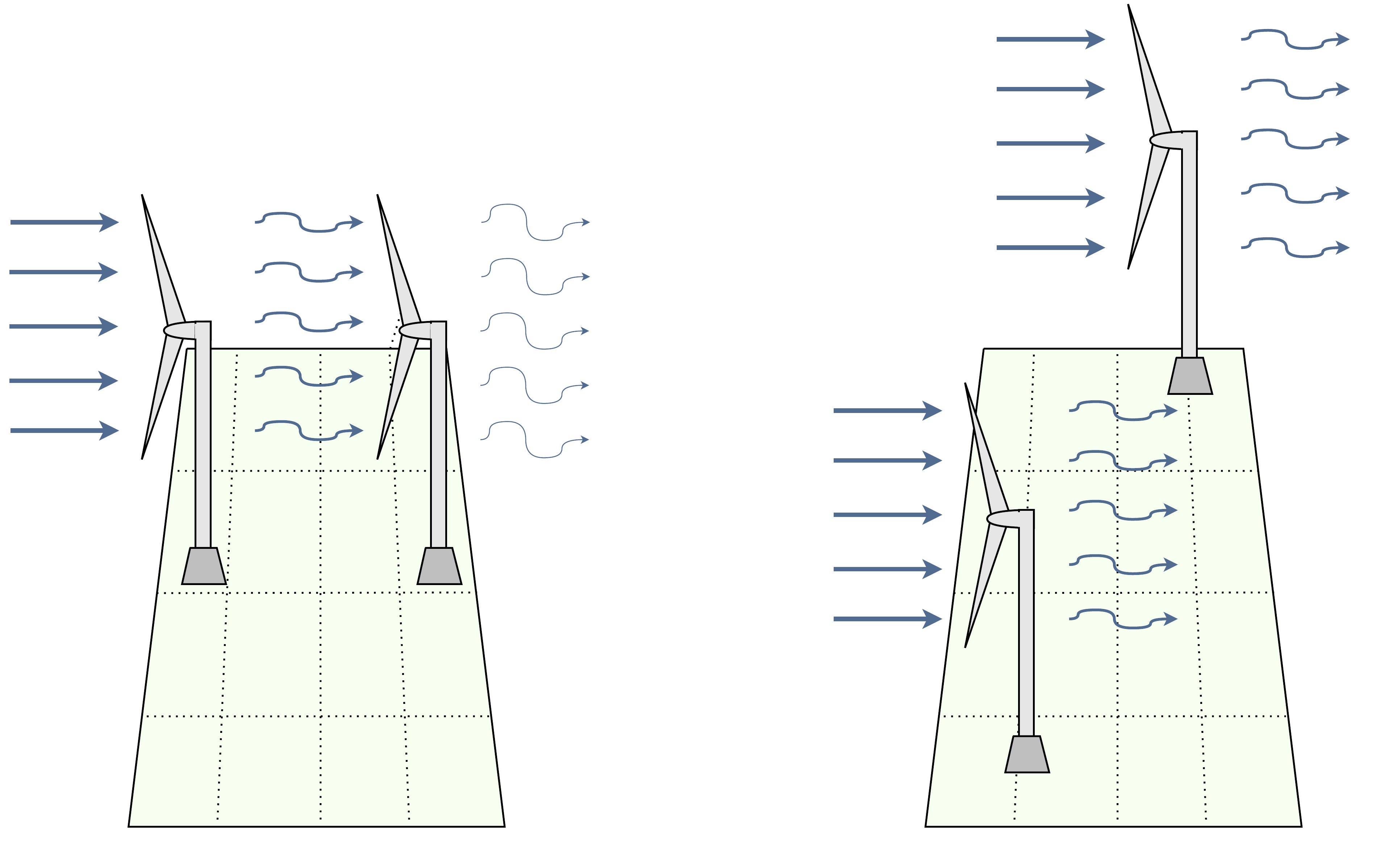}
 \caption{Planning of wind turbines within limits shown in green. The wind is shown by the blue arrows: wider arrows indicate stronger wind and more undulating arrows indicate more turbulence. In the layout on the left one turbine is directly downwind of the other, resulting in weaker and more turbulent wind. To the right, both turbines get unhindered wind.}
\label{fig:windfarm-app}
\end{figure}

\begin{figure}[h]
\centering
 \includegraphics[width=0.8\linewidth]{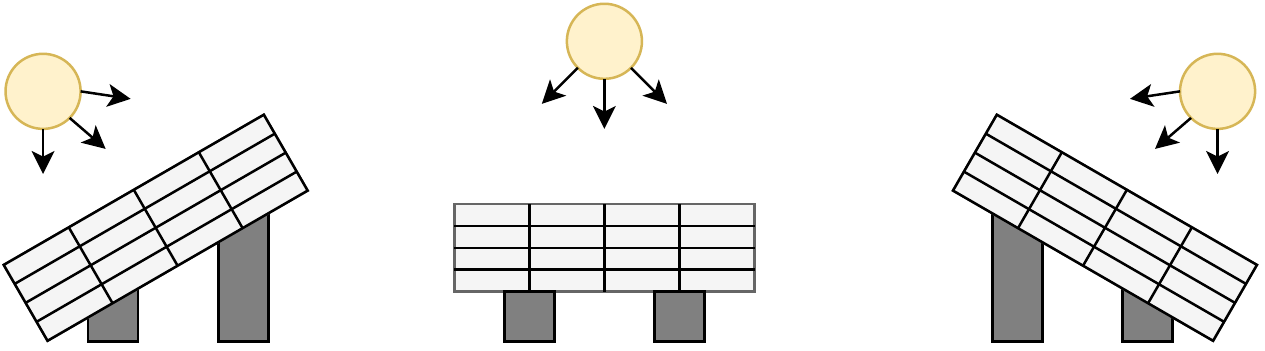}
 \caption{Adjusting solar panels to optimise power generation as the sun moves across the sky. The benchmark optimises the sum of direct, diffuse and reflective sunlight.}
\label{fig:solar-app}
\end{figure}

\begin{figure}[h!]
\centering
 \includegraphics[width=\linewidth,trim={1cm 3cm 1cm 3cm},clip]{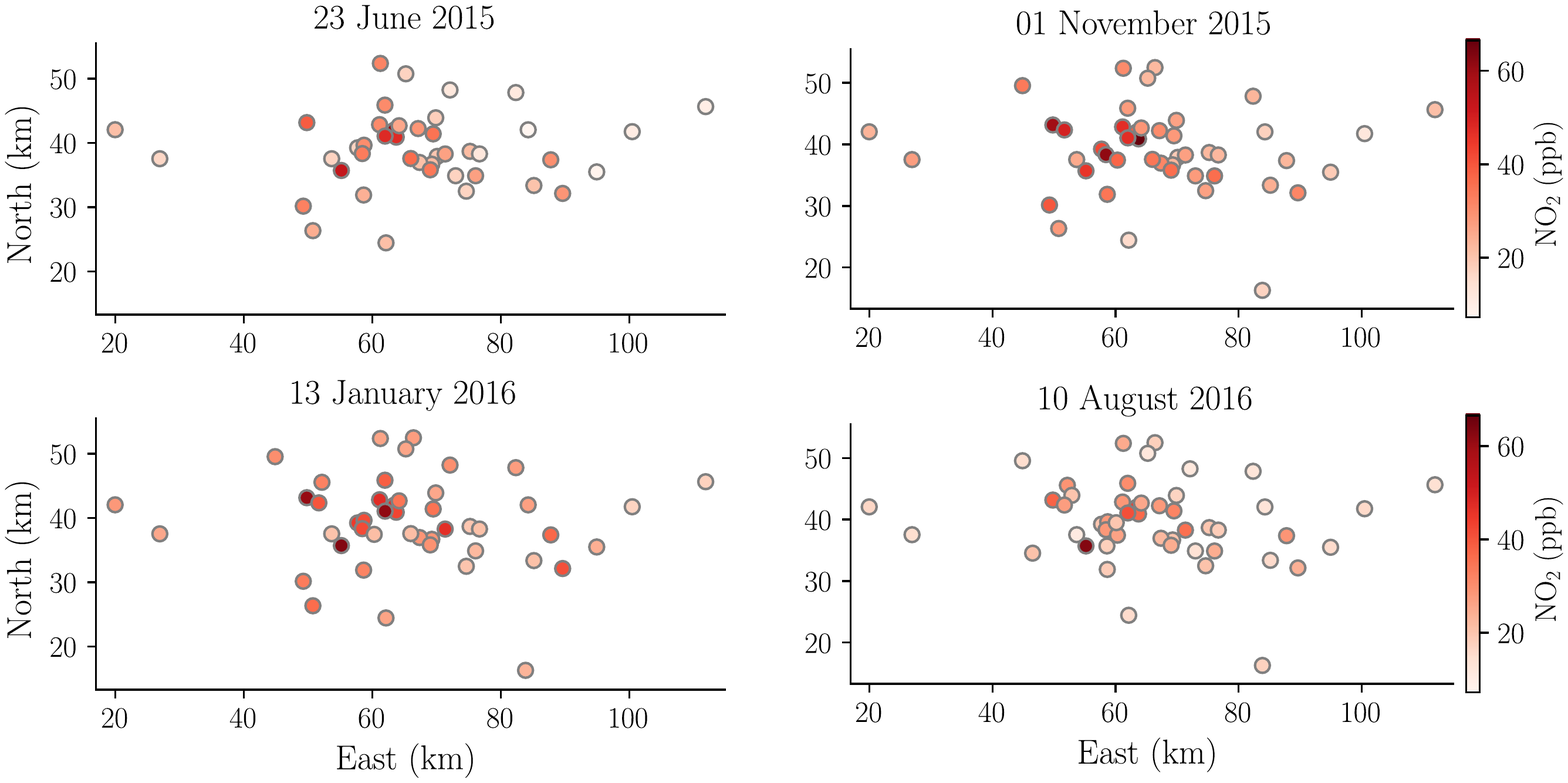}
 \caption{Example LAQN-BO problems from the training set (2015, top) and test set (2016, bottom). Not all sensors are available on every day. The overall pollution level varies, compare the top left and right plots. The locations of maxima also vary, compare the right top and bottom plots. Most of the sensors are clustered together in central London.}
\label{fig:env-examples-app}
\end{figure}

\cref{fig:windfarm-app} shows the wind farm layout problem, where the task is to place wind turbines within a permitted area. In the benchmark five turbines are placed simultaneously, here we illustrate the problem with two turbines.
\cref{fig:solar-app} illustrates the optimal renewable control benchmark. At a simple level, the problem of adjusting angles can be understood as maximising the amount of sunlight hitting the panels, which means the optimal angle varies throughout the day. The benchmark takes the challenge a step further by not only optimising the amount of \textit{direct} sunlight, but also diffusive and reflected sunlight. Our example implementation for the solar panel benchmark is discussed in \cref{sec-app-solar_rl}.
\cref{fig:env-examples-app} shows four example problems from the LAQN-BO benchmark.

\newpage

\section{Optimal renewable control implementation} \label{sec-app-solar_rl}

Code is available at \url{https://github.com/sighellan/solar_panels_rl}.

The benchmark \cite{abel2018bandit} relies on the simple\_rl Python package \citeappend{abel2019simple_rl}; we used version 0.784.
For our example Bayesian optimisation implementation we additionally relied on BoTorch \citeappend{balandat2020botorch}.
We also provide a requirements file and port the original code to Python 3. 
The implementation is not meant to be competitive with the existing methods for the benchmark, rather it is meant to demonstrate how BO can be applied.

\bibliographystyleappend{splncs04}
\bibliographyappend{short}

\end{document}